\documentclass[letterpaper]{article} 
\usepackage{aaai25}  
\usepackage{times}  
\usepackage{helvet}  
\usepackage{courier}  
\usepackage[hyphens]{url}  
\usepackage{graphicx} 
\usepackage{natbib}  
\usepackage{caption} 
\usepackage{algorithm}
\usepackage{algpseudocode}
\usepackage{color}
\usepackage{colortbl}
\usepackage{multirow}
\usepackage{tcolorbox}
 \usepackage{amsmath} 
\usepackage{booktabs}
\usepackage{newfloat}
\usepackage{listings}
\usepackage{subcaption}
\usepackage{caption}
\newcommand{\rotbox}[1]{\rotatebox{55}{#1}}

\newlength\savewidth
\newcommand{\tablestyle}[2]{\setlength{\tabcolsep}{#1}\renewcommand{\arraystretch}{#2}\centering\footnotesize}

\DeclareCaptionStyle{ruled}{labelfont=normalfont,labelsep=colon,strut=off}
\floatstyle{ruled}
\newfloat{listing}{tb}{lst}{}
\floatname{listing}{Listing}
\title{Prompt Compression with Context-Aware Sentence Encoding for Fast and Improved LLM Inference}

\author{
    Barys Liskavets\textsuperscript{1}\thanks{Corresponding Author. Email: liskovets.borets@gmail.com}, Maxim Ushakov\textsuperscript{1}, Shuvendu Roy\textsuperscript{2}, Mark Klibanov\textsuperscript{3}, Ali Etemad\textsuperscript{2}, Shane Luke\textsuperscript{3}\\
}
\affiliations{
    \textsuperscript{\rm 1}Alterra AI, Palo Alto, United States\\
    \textsuperscript{\rm 2}Queen's University, Canada\\
\textsuperscript{\rm 3}Workday Inc.
}

\usepackage{bibentry}

\begin{document}

\maketitle

\begin{abstract}
Large language models (LLMs) have triggered a new stream of research focusing on compressing the context length to reduce the computational cost while ensuring the retention of helpful information for LLMs to answer the given question. Token-based removal methods are one of the most prominent approaches in this direction, but risk losing the semantics of the context caused by intermediate token removal, especially under high compression ratios, while also facing challenges in computational efficiency. In this work, we propose context-aware prompt compression (CPC), a sentence-level prompt compression technique where its key innovation is a novel context-aware sentence encoder that provides a relevance score for each sentence for a given question. To train this encoder, we generate a new dataset consisting of questions, positives, and negative pairs where positives are sentences relevant to the question, while negatives are irrelevant context sentences. We train the encoder in a contrastive setup to learn context-aware sentence representations. Our method considerably outperforms prior works on prompt compression on benchmark datasets and is up to 10.93$\times$ faster at inference compared to the best token-level compression method. We also find better improvement for shorter length constraints in most benchmarks, showing the effectiveness of our proposed solution in the compression of relevant information in a shorter context. Finally, we release the code and the dataset for quick reproducibility and further development: https://github.com/Workday/cpc.
\end{abstract}

\section{Introduction}
The advent of large language models (LLMs) has triggered a surge of research into prompting techniques, including chain-of-thought \cite{wei2022chain}, in-context learning \cite{dong2022survey}, and retrieval augmented generation \cite{lewis2020retrieval}, aimed at leveraging their generalization and reasoning capabilities for various downstream applications. In practice, providing suitable and descriptive (often long) prompts is essential for LLMs to generate useful responses for domain-specific tasks. However, longer prompts come at a significant inference expense for LLMs in terms of both time and cost. Therefore, striking a balance between the quality and the length of the prompt is a timely area of research in order to optimize inference performance vs. cost. 

\begin{figure}[t]
    \centering
    \includegraphics[width=1.0\linewidth]{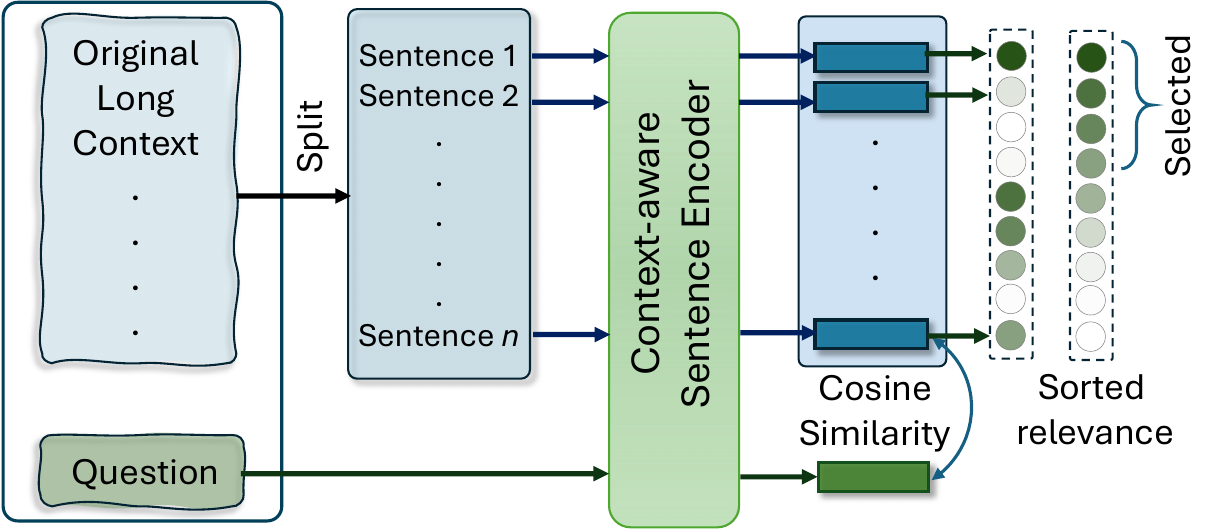}
    \caption{Overview of CPC. We first train a novel \textbf{context-aware} sentence encoder contrastively on a newly generated dataset of questions, positives, and negative pairs. Our model then uses this context-aware encoder to generate embeddings of the question and context sentences, and uses the similarity between them to select relevant sentences. }
    \label{fig:overview}
\end{figure}

\begin{table}[t]
    \centering
    \resizebox{0.72\linewidth}{!}{    
    \begin{tabular}{l c c}
    \toprule
        Method & Performance & Latency\\ \midrule
        LLMLingua & 37.2 & 9.8$\times$\\
        LLMLingua-2 & 42.2 & 0.67$\times$\\
        LongLLMLingua & 48.8 & 10.93$\times$\\
        CPC & \textbf{50.0} & 1$\times$\\ 
        \bottomrule
    \end{tabular}
}
    \caption{Overall comparison of our method on performance and latency on LongBench. CPC outperforms LongLLMLingua (SOTA) on both performance and latency.}
    \label{tab:latency}
\end{table}

Recently, prompt compression has shown promise as a solution to the context length problem. The basic idea of such a method is to reduce the size of the prompt by removing less informative content. For instance, LLMLingua \cite{jiangllmlingua} proposed to utilize a well-trained language model to identify and remove non-essential tokens from the prompt. LongLLMLingua \cite{jiang2023longllmlingua} enhanced LLMLingua for longer context by enabling question-aware compression. Later, LLMLingua-2 \cite{pan2024llmlingua} proposed to formulate prompt compression as a token-level binary classification task that learns a task-agnostic prompt compression model by removing less important tokens. However, such token removal methods may result in non-coherent sentences, often hampering the semantics of the prompt, especially when the compression ratio is relatively high. This results in a drop in the performance of the LLMs in answering questions. Additionally, most existing compression methods are usually computationally expensive as the inference time is proportional to the token length.

In this work, we propose a sentence-level compression technique called Context-aware Prompt Compression (CPC) that compresses the prompt by removing the sentences that are less relevant to the given question. A key innovation of our proposed solution is a context-aware sentence encoder that is used to rank all the sentences in the context based on their relevance to the question. Here, the relevance is measured as the (context-aware) embedding similarity (cosine distance) between the question and each sentence in the context. An important characteristic of our method is that it performs context compression while preserving human readability, unlike token-based compressors such as the LLMLingua family \cite{pan2024llmlingua, jiang2023longllmlingua, jiangllmlingua}.
We train the context-aware sentence encoder by learning to distinguish between positive and negative sentences, where the positives are context sentences that contain relevant information to the given question, and the negatives are context sentences that do not contain any relevant information to the question. Along with the proposed method, we also introduce a new dataset with question, positive, and negative sentence pairs required to train our context-aware sentence encoder. 
Figure \ref{fig:overview} provides an overview of our method.

We evaluate our model following the protocol established by prior works on prompt compression \cite{pan2024llmlingua, jiang2023longllmlingua} on LongBench \cite{bai2023longbench} and ZeroSCROLLS \cite{shaham2023zeroscrolls}. Extensive experiments show that our proposed method outperforms the existing state-of-the-art (LongLLMLingua \cite{jiang2023longllmlingua}) on these two benchmarks by 1.5\% and 1.3\% on average on the 2,000 tokens constraint. Similarly, on the 3,000 token constraint, our method outperforms others by 1.2\% and 2.1\%, respectively. On the individual sub-tasks of LongBench, our method shows up to 8.3\% improvements. 
We present detailed ablation studies on different components of our proposed solution, including the dataset collection and the context-aware sentence embedding module. Furthermore, our method is up to 10.93$\times$ faster than LongLLMLingua during inference, imposing minor overhead during prompt compression (see Table \ref{tab:latency}). 
In summary, our contributions are:
\begin{itemize}
    \item We propose a new method for sentence-level prompt compression. Our method relies on the novel concept of context-aware sentence encoding, which enables it to compress prompts by removing sentences that do not contain relevant information for the given question. 
    \item As part of our solution, we curate and release a new dataset that contains tuples of context, question, positive, and negatives to train the context-aware sentence encoder. 
    \item Our method outperforms prior works and sets a new state-of-the-art for prompt compression on LongBench and ZeroSCROLLS. Additionally, our method is up to 10.93$\times$ faster during inference compared to existing prompt compressors. To enable reproducibility and contribute to the area, we release our code and dataset at: https://github.com/Workday/cpc. 
\end{itemize}

\section{Related work}
In this section, we discuss the literature relevant to our work from two perspectives. First, we review previous works on prompt compression. Next, we provide an overview of the literature on sentence embedding learning, focusing on aspects relevant to our context-aware embedding module.

\subsection{Prompt compression.}
Recent work on prompt compression aims to reduce the inference cost of LLMs. A prominent direction in this area is model-agnostic prompt compression, which utilizes a pre-trained model to act as an information-based compressor. For instance, \citet{kim2022learned} introduced a token pruning method in the forward pass of the LLM. In \citet{chenwalking}, a heuristic method was employed to recursively summarize the context. A limitation of these early works is that they require access to the pre-trained LLM as part of the compression method, which is not often feasible. Some of the more recent works developed solutions that do not require the LLM for compression. The most notable method in this direction is LLMLingua \cite{jiangllmlingua}, which uses token-level perplexities to filter out semantically insignificant tokens that do not have high perplexity. Similar efforts have been performed by Selective-Context \cite{li2023compressing}, with decoder-only models (LLaMa \cite{touvronllama}, GPT-2 \cite{gpt_2}) for token-wise perplexity calculation, which keeps a pre-defined number of highest-perplexity tokens to achieve the required compression. LongLLMLingua \cite{jiang2023longllmlingua} further developed this idea for question-aware context compression by adding document-level question relevance estimation prior to employing LLMLingua. In general, these methods suffer from the lack of capability to adapt to new domains. 

Another common direction of prompt compression is trainable prompt compression methods, including soft prompt methods, sequence-to-sequence methods, and reinforcement learning methods, among others. Soft prompt methods directly pre-train or fine-tune a language model as the compressor \citep{wang2024adapting,bulatov2022recurrent, chevalier2023adapting, gecontext, wang2024loma}, which usually yield a high compression rate but have little interpretability or control over the compression rate. Sequence-to-sequence models capture the entire context as input and directly generate the compressed sentence \cite{xu2024recomp}. However, these methods have high latency due to the autoregressive nature of generation. Reinforcement learning for prompt compression, such as in \citet{labankeep}, introduces a novel reward function to simplify complex text by jointly optimizing simplicity, fluency, salience, and guardrails. Another approach utilizes compression ratio as a reward function, conditioned on meeting a ROUGE metric threshold \cite{jung2024discrete}, but may compromise on question-aware context compression tasks. Furthermore, reinforcement learning has been employed for token classification in efficient unsupervised sentence compression by fine-tuning pre-trained encoder \cite{ghalandariefficient}. Among more recent methods, \citet{pan2024llmlingua} trains a model that is designed to evaluate the information value of each specific lexical unit, which is then sorted by these values and pruned. However, the token-level compression results in a non-coherent sentence that compromises the semantics of the compressed prompt, resulting in a sub-optimal performance of the LLM. In this work, we focused on prompt compression by identifying and removing less relevant \textit{sentences} from an input context.

\subsection{Text embedding learning.}
The goal of text embedding learning is to generate text embeddings that capture the semantic meaning in a high-dimensional vector space, enabling various natural language processing tasks, such as text classification, clustering and similarity search. Earlier works in text embedding learning, such as GloVe \cite{pennington2014glove} and Word2Vec \cite{church2017word2vec}, focused on learning a word or token-level representation, while more recent works have explored sentence-level representation learning. For instance, \citet{reimers2019sentence} proposed to fine-tune BERT-like architecture \cite{vaswani2017attention} for extraction of sentence embeddings that can be subsequently utilized to measure text similarities. Later \citet{li2023towards,wang2022text,beltagy2020longformer} further improved the efficacy of the text embedding, especially for longer contexts. More recently, \citet{behnamghader2024llm2vec} utilized the vast knowledge of pre-trained LLMs to develop a strong sentence encoder. However, the sentence representation learned by such models is not context-aware, and no prior works focused on context-aware sentence encoding that is required for our prompt compression approach.

\section{Method}

\subsection{Problem definition}
Let $x=\{{x_i}\}_{i=1}^L$ be an input context of length $L$ tokens. The goal of the prompt compression is to produce a compressed prompt $\tilde{x}=\{\tilde{x}_{i}\}_{i=1}^{\tilde{L}}$, where $\tilde{L}<L$. The compression ratio can be denoted as $\tau = \tilde{L}/L$, where $\tau \in [0,1]$. The goal of an efficient compression method is to generate a compressed prompt (with smaller $\tau$) while keeping the relevant context preserved so that the performance of the LLM on the compressed prompt~$\tilde{x}$ matches that of the original prompt~$x$. However, existing solutions often rely on token-level compression, which can come with a key drawback. The removal of intermediate tokens from a sentence may result in a non-coherent and grammatically incorrect sentence, often hampering the semantics of the input prompt and causing a drop in the performance of LLM. In this work, we propose a sentence-level compression method that removes sentences from the given context based on their relevance to the input question. The key innovation of our proposed method is a context-aware sentence encoder, which is utilized to find the relevance of each sentence in the context, given a question. Our method requires a dataset of context, question, positive, and negative tuples for training our proposed context-aware sentence encoder, which we discuss below. Subsequently, we discuss our proposed method for training the context-aware sentence encoder and prompt compression pipeline during inference.

\subsection{Dataset curation}
To train the context-aware sentence encoder, we first create a dataset containing tuples of (long) contexts, questions, positive sentences, and negative sentences, called Context-aware Question-Relevance (CQR) Dataset.
The `context' is the text input that contains all the relevant information to answer the `question', which is a meaningful query that asks some key information regarding the `context'. 
The `positives' are defined as sentences within the `context' that contain \textit{some}, but not necessarily all relevant information to answer the question. Finally, the `negatives' are `context' sentences that contain no information relevant to answering the question. To generate a dataset of such tuples, we use a two-step approach where we first generate a set of questions, positives, and negatives, followed by a filtering step. We discuss these two steps below.

We start with the WikiText dataset \cite{merity2016pointer} as the seed dataset for our synthetic data generation. This dataset consists of long Wikipedia pages, each containing factual information about historical/scientific concepts. First, we take a document from the original dataset which we consider as context $C$. This document consists of sentences~$\{S_i\}_{i=1}^K$. Next, we sample `positive' sentence $P$ from $S_i$. During the sampling procedure, we ensure that $P$ is a consistent English sentence that contains coherent information.
We consider a sentence to be consistent and coherent if at least $\theta\%$ of its words are from the English dictionary and consist of ASCII-only characters.
Then, we prompt a pre-trained LLM $\psi$ to generate synthetic question-answer pairs $(Q_j, A_j)$ for this particular sentence $P$ while also considering its context $C$. This is performed using Prompt 1 below.
\noindent\fbox{\begin{minipage}{23.5em}
{\small
\noindent \texttt{\textbf{Prompt 1 (Question Prompt):}\\
Here is a text to consider: TEXT: "{text}"\\
Read the sentence in double brackets, namely, [[{sentence}]].\\
Ask questions to this sentence, and make sure the question is not answerable from this sentence alone without knowing the
context.\\
Reply in this format:\\
\textbf{Q:} \{question 1\}\\
\textbf{A:} \{answer 1\}\\
\textbf{Q:} \{question 2\}\\
\textbf{A:} \{answer 2\}}
}

\end{minipage}}
\\\\

Note that while some sentences may not directly contain the information required to answer the question, they may provide key contextual clues needed to do so indirectly. For example:

\noindent\fbox{\begin{minipage}{23.5em}
{\small
\noindent \texttt{Q: How many children does John have?\\
C: (1) John and Mary have been married for 10 years. (2) They have their anniversary now and they have decided to travel to Spain for a month. (3) Mary is worried that her two children are too small to travel on the plane.(4) So she has asked her sister to look after them.}}
\end{minipage}}
~

In the example above, sentences (1) and (3) do not contain sufficient information to answer question $Q$, but together they form a subset of sentences that contains all the necessary information. Note that if we evaluate the sentence embedding similarity of the question and each of these sentences, the similarity between sentence (3) and $Q$ would be relatively low due to the absence of any direct reference to John.
Accordingly, we need our compression model to train on questions that can not be answerable solely based on $P$.
To ensure the question is not fully answerable from the sentence $P$ alone, we use the LLM to verify each sentence/question/answer triplet with Prompt 2 as follows:

\noindent\fbox{\begin{minipage}{23.5em}
{\small
\noindent \texttt{\textbf{Prompt 2 (Verification Prompt):}\\
You are given a piece of text, a question and an answer. Verify whether it is possible to derive such an answer by considering only the given piece of text (you should rely only on the piece of text). Think step by step and finish your thoughts with one word: "Yes" or "No". Answer "Yes" if and only if ALL the necessary information is contained in the text. If anything is missing, then state what is missing and answer "No". Answer "Yes" ONLY if there is no such information in the answer that is missing in the text. Otherwise, answer "No"!!\\
\{A demonstration\}\\
Text: \{context sentence\}\\
Question: \{question\}\\
Answer: \{answer\}\\
Verification result: Yes/No}
}
\end{minipage}}
~\\

Here the $Q\text{\&}A$ generated pairs that have passed the verification step (obtained verification result ``No'') along with $P$ and $C$ are kept in the dataset. 
Next, to gather the negative sentences, we employ a pre-trained sentence transformer \cite{reimers2019sentence}, denoted as $\mathcal{E}$. This transformer generates embeddings for each sentence $S_i$ in the context, represented as $\{E_i=\mathcal{E}(S_i)\}_{i=1}^{K}$. Additionally, it produces embeddings for the question $Q$ and the positive sentence $P$, denoted as $E_q=\mathcal{E}(Q)$ and $E_p=\mathcal{E}(P)$, respectively. We define a similarity score $\eta$ as the cosine similarity between the embeddings of the positive sentence and the question, $\eta = \cos(E_p, E_q)$, which we use to identify candidate negative sentences from the context. These candidates are the sentences whose similarity to the question is less than $\eta$, i.e., $\cos(\mathcal{E}(S_j), E_q) < \eta$ for any sentence $S_j$ in the context. In case the number of sentences with a similarity score less than or equal to $\eta$ is less than $\beta\%$ of the total sentences, i.e. $\left|\{i \mid \cos(E_i, E_q) \le \eta\}\right| < \beta \cdot K$, we exclude such ($C$, $Q$, $P$) tuples from our dataset.

\begin{figure*}[t]
    \centering
    \includegraphics[width=1.0\linewidth]{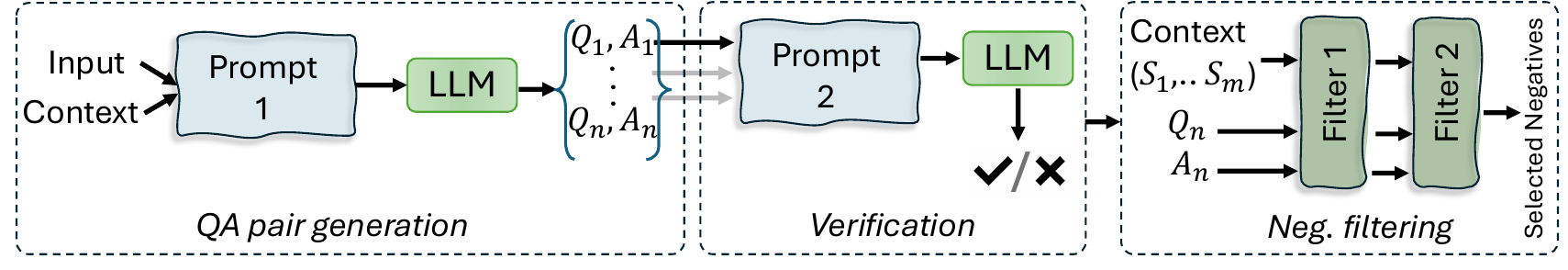}
    \caption{Data collection pipeline. The data collection process starts by asking the LLM with \textit{Prompt 1} to generate a set of ($Q$,$A$) pairs. In the next step, we verify the pairs with \textit{Prompt 2}. In the last step, we select a set of negatives for each ($Q$,$A$) pair.}
    \label{fig:dataset}
    \vspace{10pt}
\end{figure*}

Once we generate a set of negative candidates, we introduce a negative filtering method using a probability density function to further filter the negatives. Specifically, for a negative sentence $S_{Neg}$, we use a pre-trained LLM, $\phi$, to obtain the probability density of the answer tokens $A$, given the context $C$ without the negative sentence as:
\begin{equation}
  P_{C\ \text{w/o}\ Neg}=\phi\left(A \mid \{S_{i}\}_{i=1..K, i\ne neg}\right).
\end{equation}
Similarly, we calculate the probability density of the answer tokens, given the whole context as:
\begin{equation}
  P_{C}=\phi\left(A \mid \{S_{i}\}_{i=1..K}\right).
\end{equation}
For a negative sentence with no information about the question, these two distributions should be close. If the KL-divergence between two distributions ($KL(P_{C},P_{C\ \text{w/o}\ Neg}$) is high ($>\lambda$), it indicates that the negative sentence has information regarding the question. Consequently, we filter out such negatives from our dataset. An overview of the data collection pipeline is illustrated in Figure \ref{fig:dataset}.

\subsection{Context-aware sentence encoding}
Next, our pipeline relies on a context-aware sentence encoder, $f_\theta$ to capture the semantics of a sentence or question, given its context. 
We train $f_\theta$ on our \textit{CQR dataset} in a contrastive learning setup. Specifically, $f_\theta$ is trained to learn the context-aware representation by minimizing distances between the question and positive samples (context sentences that contain some information regarding the question) while maximizing it from the negative samples (irrelevant context sentence). Let's denote a training batch of our dataset be, $\mathcal{B}=\{C_{b}, Q_{b}, P_{b}, \{N_{b}^{n=1..M}\}\}_{b=1..B}$, where $C$, $Q$, $P$, and $N$ are the context, question, positive, and negatives. For a sample $1 \leq b \leq B$, we consider the exponent of the cosine similarity between the question's semantic embedding ($\xi_{Q_{b}}$) and the context-aware sentence embedding of the positive sample ($\xi_{(P_{b}, C_{b})}$) as positives, i.e, $Sim_{P}=exp(cosine(\xi_{Q_{b}}, \xi_{(P_{b}, C_{b})}))$, where $\xi_{P,C}$ is the context-aware representation of sentence $P$ in the context of $C$, defined as:
\begin{equation}
    \xi_{P,C}=norm(\frac{\sum_{t=i}^j Z_t }{|j-i+1|}).
\end{equation}
Here, $Z_i$ is the token-level embedding of token $x_i$,  calculated as: 
\begin{equation}
    \{Z_{i}\}^{i=1..L} = f_\theta(x_0, ..., x_L),
\end{equation}
where $i$ and $j$ are the start and end indexes of the positive sentence $P$ in context $C$. Similarly, the question embedding (without the context) is represented as $\xi_{Q}$. Similarly, we define the negative as $Sim_{N}=exp(cosine(\xi_{Q_{b}}, Neg_{(b,ext)})$, where $Neg_{(b,ext)}$ consists of the corresponding negative sentences of sample $b$, along with all the positives and negatives from the same batch $\mathcal{B}$ excluding $b$, i.e. 
$Neg_{(b, ext)}=\{\xi_{(P_{i}, C_{i})}\}_{i=1..B | i\neq b} \cup \{\xi_{(N_{i}^n, C_{i})}\}_{i=1..B | i\neq b}^{n=1..M} \cup \{\xi_{(N_b^n, C_b)}\}^{n=1..M} $. 
Consequently, we denote our contrastive training loss as:
\begin{equation}
    \mathcal{L}_{SC} = -log \frac{exp(Sim_P)}{exp(Sim_P) + \sum exp(Sim_{N})},
\end{equation}

In practice, we do not train $f_\theta$ from scratch. Rather, we adopt a pre-trained LLM and use the concept introduced in \citet{behnamghader2024llm2vec} through finetuning to learn bidirectional attention over the whole context. More specifically, we mask $\delta$\% of the tokens from the context, and train the model with a masked next token prediction loss:
\begin{equation}
     \mathcal{L}_{MNTP} = \sum_{i}^{tokenlen(C)} -logP(x_i | X_{um}),
\end{equation}
where $X_{um}$ is the sequence of all the unmasked tokens and $P(x_i)$ is the probability distribution over the vocabulary for the $i$th masked token. Finally, we train the model with the combination of the two losses: $\mathcal{L} = \mathcal{L}_{SC} + \mathcal{L}_{MNTP}$.

\begin{table*}[tb]
    \centering
    \setlength{\tabcolsep}{1mm}
    \begin{tabular}{l|ccccccccc|ccc}
    \toprule
        \multirow{2}{*}{\textbf{Methods}} &  \multicolumn{9}{@{}c}{{\bf LongBench}} &  \multicolumn{3}{@{}c}{{\bf ZeroSCROLLS}} \\
        & SingleDoc & MultiDoc & Summ. & FewShot & Synth. & Code & AVG & Tokens & $1/\tau$ & AVG & Tokens & $1/\tau$\\
    \midrule
    \midrule
        Original Prompt & 39.7 & 38.7 & 26.5 & 67.0 & 37.8 & 54.2 & 44.0 & 10,295 & - & 32.5 & 9,788 & -  \\
    \cmidrule (r){1-1}\cmidrule (lr){2-10} \cmidrule (lr){11-13}
    Zero-shot & 15.6 & 31.3 & 15.6 & 40.7 & 1.6 & 36.2 & 23.5 & 214 & 48$\times$ & 10.8 & 32 & 306$\times$ \\
    \midrule
    \midrule
    \multicolumn{6}{@{}r}{{ \textit{3,000 tokens constraint}}} \\
    \midrule
      \multicolumn{13}{@{}l}{{ \textit{Retrieval-based Methods}}} \\ 
    BM25 & 32.3 & 34.3 & 25.3 & 57.9 & 45.1 & 48.9 & 40.6 & 3,417 & 3$\times$ & 19.8 & 3,379 & 3$\times$ \\
    SBERT & 35.3 & 37.4 & 26.7 & 63.4 & 51.0 & 34.5 & 41.4 & 3,399 & 3$\times$ & 24.0 & 3,340 & 3$\times$\\
    OpenAI & 34.5 & 38.6 & 26.8 & 63.4 & 49.6 & 37.6 & 41.7 & 3,421 & 3$\times$ & 22.4 & 3,362 & 3$\times$  \\
    LongLLMLingua & 37.6 & 42.9 & 26.9 & 68.2 & 49.9 & 53.4 & 46.5 & 3,424 & 3$\times$ & 29.3 & 3,350 & 3$\times$ \\
    \cmidrule (r){1-1}\cmidrule (lr){2-10} \cmidrule (lr){11-13}
    \multicolumn{7}{@{}l}{{ \textit{Compression-based Methods}}} \\
    Selective-Context & 23.3 & 39.2 & 25.0 & 23.8 & 27.5 & 53.1 & 32.0 & 3,328 & 3$\times$ & 20.7 & 3,460 & 3$\times$ \\
    LLMLingua & 31.8 & 37.5 & 26.2 & 67.2 & 8.3 & 53.2 & 37.4 & 3,421 & 3$\times$ & 30.7 & 3,366 & 3$\times$ \\
    {LLMLingua-2} &  35.5 & 38.7 & 26.3 & 69.6 & 21.4 & 62.8 & 42.2 & 3,392 & 3$\times$ & 33.5 & 3,206 & 3$\times$ \\
    {LongLLMLingua} &  40.7 & 46.2 & \textbf{27.2} & \textbf{70.6} & \textbf{53.0} & {55.2} & 48.8 & 3,283 & 3$\times$ & 32.8 & 3,412 & 3$\times$ \\
    \cmidrule (r){1-1}\cmidrule (lr){2-10} \cmidrule (lr){11-13}
    {\cellcolor[rgb]{0.925,0.957,1}}\textbf{CPC (Ours)} &    {\cellcolor[rgb]{0.925,0.957,1}}\textbf{42.7} & {\cellcolor[rgb]{0.925,0.957,1}}\textbf{47.9} & {\cellcolor[rgb]{0.925,0.957,1}}23.8 & {\cellcolor[rgb]{0.925,0.957,1}}69.3 & {\cellcolor[rgb]{0.925,0.957,1}}52.8 & {\cellcolor[rgb]{0.925,0.957,1}}\textbf{63.5} & {\cellcolor[rgb]{0.925,0.957,1}}\textbf{50.0} & {\cellcolor[rgb]{0.925,0.957,1}}3,327 & {\cellcolor[rgb]{0.925,0.957,1}}3$\times$ & {\cellcolor[rgb]{0.925,0.957,1}}\textbf{34.9} & {\cellcolor[rgb]{0.925,0.957,1}}3,470 & {\cellcolor[rgb]{0.925,0.957,1}}3$\times$\\
    \midrule
    \midrule
    \multicolumn{6}{@{}r}{{ \textit{2,000 tokens constraint}}} \\
    \midrule
      \multicolumn{13}{@{}l}{{ \textit{Retrieval-based Methods}}} \\ 
    BM25 & 30.1 & 29.4 & 21.2 & 19.5 & 12.4 & 29.1 & 23.6 & 1,985 & 5$\times$ & 20.1 & 1,799 & 5$\times$\\
    SBERT & 33.8 & 35.9 & 25.9 & 23.5 & 18.0 & 17.8 & 25.8 & 1,947 & 5$\times$ & 20.5 & 1,773 & 6$\times$ \\
    OpenAI & 34.3 & 36.3 & 24.7 & 32.4 & 26.3 & 24.8 & 29.8 & 1,991 & 5$\times$ & 20.6 & 1,784 & 5$\times$ \\
    LongLLMLingua & 37.8 & 41.7 & 26.9 & 66.3 & 53.0 & 52.4 & 46.3 & 1,960 & 5$\times$ & 24.9 & 1,771 & 6$\times$ \\
    \cmidrule (r){1-1}\cmidrule (lr){2-10} \cmidrule (lr){11-13}
    \multicolumn{7}{@{}l}{{ \textit{Compression-based Methods}}} \\
    Selective-Context & 16.2 & 34.8 & 24.4 & 15.7 & 8.4 & 49.2 & 24.8 & 1,925 & 5$\times$ & 19.4 & 1,865 & 5$\times$ \\
    LLMLingua & 22.4 & 32.1 & 24.5 & 61.2 & 10.4 & {56.8} & 34.6 & 1,950 & 5$\times$  & 27.2 & 1,862 & 5$\times$ \\
    LLMLingua-2 & 29.8 & 33.1 & 25.3 & 66.4 & 21.3 & 58.9 & 39.1 & 1,954 & 5$\times$ & 33.4 & 1,898 & 5$\times$ \\
    {LongLLMLingua} & 39.0 & 42.2 & \textbf{27.4} & 69.3 & \textbf{53.8} & 56.6 & 48.0 & 1,809 & 6$\times$ & 32.5 & 1,753 & 6$\times$\\
    \cmidrule (r){1-1}\cmidrule (lr){2-10} \cmidrule (lr){11-13}
    {\cellcolor[rgb]{0.925,0.957,1}}\textbf{CPC (Ours)} & {\cellcolor[rgb]{0.925,0.957,1}}\textbf{42.6} & {\cellcolor[rgb]{0.925,0.957,1}}\textbf{48.6} & {\cellcolor[rgb]{0.925,0.957,1}}23.7 & {\cellcolor[rgb]{0.925,0.957,1}}\textbf{69.4} & {\cellcolor[rgb]{0.925,0.957,1}}52.8 & {\cellcolor[rgb]{0.925,0.957,1}}\textbf{60.00} & {\cellcolor[rgb]{0.925,0.957,1}}\textbf{49.5} & {\cellcolor[rgb]{0.925,0.957,1}}1,844 & {\cellcolor[rgb]{0.925,0.957,1}}5$\times$ & {\cellcolor[rgb]{0.925,0.957,1}}\textbf{33.8} & {\cellcolor[rgb]{0.925,0.957,1}}1,821 & {\cellcolor[rgb]{0.925,0.957,1}}5$\times$\\

    \bottomrule
    \end{tabular}
    \caption{Performance of different methods under different compression ratios on LongBench and ZeroSCROLLS. 
    }
    \label{tab:main_result_long_context}
    \vspace{10pt}
\end{table*}

\subsection{Inference}
At inference, given context $C$ and question $Q$, our aim is to generate a compressed context $C_{compressed}$ that preserves all the relevant information to answer the question $Q$. We start by splitting the context into sentences $\{S_i\}_{i=1}^K$ and generating the context-aware embedding for each sentence $\xi_{S_i,C}$ as explained above. Similarly, we generate the question embedding as $\xi_{Q}$. Next, we calculate the embedding similarly between $\xi_{Q}$ and each $\xi_{S_i,C}$ to generate the relevance score for all the sentences: $S_i = cosine(\xi_{S_i, C}, \xi_{Q})$. 
For a compression ratio of $\tau$, we take the top $T$ sentences with the highest relevance scores such that $\sum_{t=1..T}tokenlen(S_{t}) \leq \tau \cdot tokenlen(C)$. We then restore the order of selected sentences in the original context to form the compressed context $C_{compressed}$.

\section{Experiments}
\subsection{Datasets}\label{sec:dataset}
We train our proposed context-aware sentence encoder on our own collected dataset (CQR dataset) described earlier. For evaluation, we follow the experimental setup of previous works \cite{jiang2023longllmlingua} and report the performance of our proposed solution on LongBench \cite{bai2023longbench} and ZeroSCROLLS \cite{shaham2023zeroscrolls}. LongBench is a dataset with different sub-tasks, including single-document QA, multi-document QA, and more. Similarly, ZeroSCROLLS is another dataset with different sub-tasks, including summarization, QA, and few-shot learning. We also evaluate on domain-specific benchmarks such as medical paper summarization (PubMed \citep{cohan2018discourse}), keyword extraction (Krapivin \citep{krapivin2009large}), meeting transcript summarization (MeetingBank \citep{hu2023meetingbank}) and TV show summarization (SummScreen \citep{chen2021summscreen}) datasets.

\begin{table}[t]
    \centering
\resizebox{1\columnwidth}{!}{    
    \setlength
\tabcolsep{2pt}
    \small
    \begin{tabular}{lccccccc}
        \toprule
        \rotbox{\textbf{Method}} & \rotbox{\textbf{Krapivin-2009}} & \rotbox{\textbf{PubMed}} & \rotbox{\textbf{MeetingBank}} & \rotbox{\textbf{SummScreen}} \\ \midrule
        
       LLMLingua-2 & 36.7 & 26.24 & 16.73 & 21.57 \\ 
        LongLLMLingua & 17.1 & 24.31 & 14.39 & 17.48 \\
        CPC & \textbf{43.5} & \textbf{27.59} & \textbf{17.66} & \textbf{21.83} \\

    \bottomrule        
    \end{tabular}
    }
    \caption{Comparison to prior works on different domain-specific evaluation tasks.}
    \label{tab:domain_gen}
\end{table}

\begin{table*}
    \centering
    \setlength
\tabcolsep{7pt}
    \begin{tabular}{lccccccc}
        \toprule
        {\textbf{Embed.}} & {\textbf{Sin. Doc}} & {\textbf{Mul. Doc}} & {\textbf{Summ.}} & {\textbf{FewShot}} & {\textbf{Synth.}} & {\textbf{Code}} & {\textbf{AVG}}\\ \midrule
        
        LLMLingua-2 & 33.39 & 52.07 & 24.88 & 46.94 & 27.0 & 59.84 & 40.69 \\ 
        LongLLMLingua & 26.70 & 39.25 & 22.66 & 58.45 & 13.75 & 27.73 & 31.42 \\
        CPC (Ours) & 42.43 & 62.86 & 25.74 & 69.25 & 53.75 & 63.44 & 52.91
 \\ 
    \bottomrule        
    \end{tabular}
    \caption{Comparison to prior works with larger encoder.}
    \label{tab:gpt4o}
    \vspace{10pt}
\end{table*}

\subsection{Implementation details}

For collecting our dataset, we use WikiText \cite{merity2016pointer} as our seed corpus. Our method is flexible towards the choice of pre-trained LLM that can be used as $\psi$ and $\phi$ in our dataset collection method. However, given that we need strong text generation capabilities for $\psi$ and access to the probability density outputs for $\phi$, we chose the OpenAI \textit{gpt-3.5-turbo-instruct} and \textit{microsoft/Phi-3-small-128k-instruct} respectively. We used \textit{sentence-transformers/all-mpnet-base-v2}  as the sentence encoder \cite{reimers2019sentence} for negative collection. Finally, we set $\beta=30\%$,  $\lambda=4e-3$, $\delta=80\%$, and $M=2$.

For $f_\theta$, we adopt the \textit{Mistral-7B-Instruct-v0.2} \cite{jiang2023mistral} given that it is open-source and allows for long context lengths, and fine-tune this model with \textit{LoRA} \cite{hu2021lora} of rank 16 to learn the context-aware embeddings. We train the model with 2048 samples with a batch size 32, and AdamW optimizer with a learning rate of $5e-5$. The training is conducted on a single A100 80G GPU.

\subsection{Evaluation protocols}
We perform various experiments on several downstream tasks based on commonly used protocols. We evaluate the effectiveness of our method on summarization problems by calculating the Rouge metric \citep{lin2004rouge} between the ground truth responses versus the model's generated outputs given compressed prompts. For document QA tasks, we measure the F1 score between the response generated by the model and the ground truth response. For code completion tasks, we use the textual similarity metric based on the Levenshtein edit distance \citep{yujian2007normalized}. For the different subsets of LongBench and ZeroSCROLLS, we follow the same evaluation protocol adopted by existing literature \cite{jiang2023longllmlingua}. For the keyword extraction, we evaluate the recall of the keywords that LLM extracts from a compressed context in relation to ground truth keywords.

\begin{table}
    \centering
    \resizebox{1.0\linewidth}{!}{
    \begin{tabular}{l c c c c }
    \toprule
        \textbf{Method} & \textbf{Avg.} & \textbf{Med.} & \textbf{Avg. Rel.} & \textbf{Med. Rel.} \\ \midrule
        LLMLingua & 4.73 & 1.47 & 16.89$\times$ & 9.8$\times$ \\
        LLMLingua-2 & 0.23 & 0.1 & 0.82$\times$ &  0.67$\times$ \\
        LongLLMLingua & 7.70 & 1.64 & 27.5$\times$\ & 10.93$\times$\\
        CPC & 0.28 & 0.15 & 1$\times$ & 1$\times$ \\ 
        \bottomrule
    \end{tabular}
    }
    \caption{Inference latency for different methods. Here, Avg., Med., and Rel., stand for average, median and relative time.}
    \label{tab:latency}
\end{table}

\begin{figure}[t]
  \centering
  \begin{subfigure}{0.49\columnwidth}
    \centering
    \includegraphics[width=\textwidth]{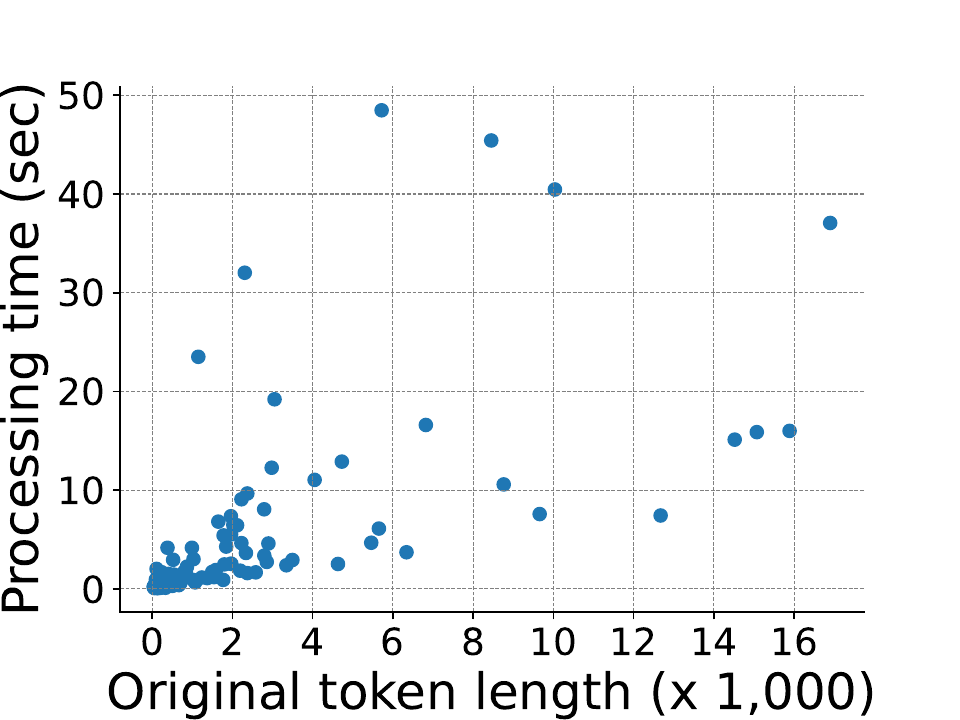}
    \caption{LongLLMLingua}
    \label{fig:image1}
  \end{subfigure}%
  \begin{subfigure}{0.49\columnwidth}
    \centering
    \includegraphics[width=\textwidth]{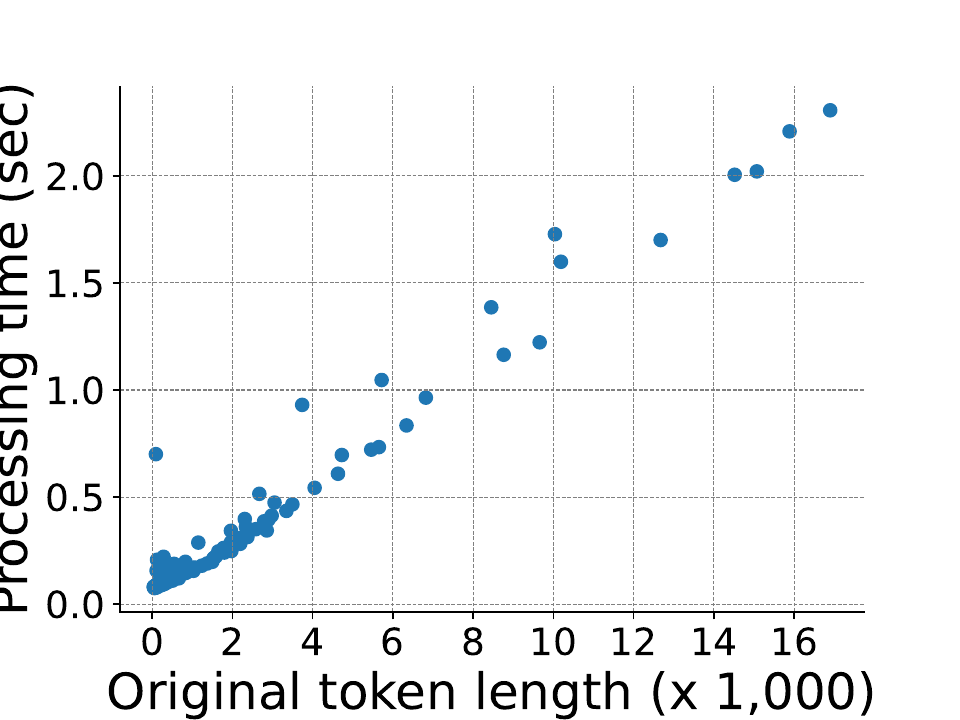}
    \caption{Ours}
    \label{fig:image2}
  \end{subfigure}
  \caption{Comparison of latency between LongLLMLingua and our proposed method.}
  \label{fig:latency}
\end{figure}

\begin{figure*}
    \centering
    \includegraphics[width=1.0\linewidth]{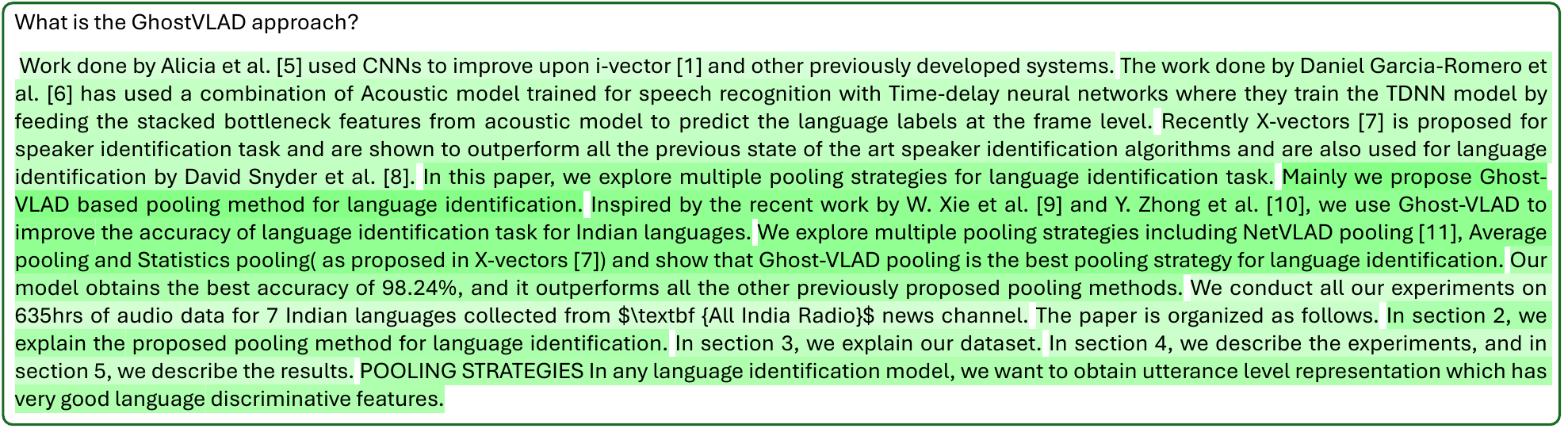}
    \caption{An example illustrating the question and sentence relevance scores from our context-aware sentence encoder. Here, the more relevant sentences are coloured with higher intensity. }
    \label{fig:color-map}
    \vspace{10pt}
\end{figure*}

\begin{table*}[t]
\resizebox{1.0\linewidth}{!}{
    \centering
    \subfloat[
     Ablation on loss function.
    \label{tab:ablation_loss}
    ]{
    \centering
    \begin{minipage}{0.24\linewidth}{\begin{center}
    \tablestyle{2pt}{1.0}
        \begin{tabular}{l c}
        \toprule
        \textbf{Ablation} & \textbf{Accuracy} \\
        \midrule
        $\mathcal{L}_{SC} +  \mathcal{L}_{MNTP}$ & \textbf{42.38} \\

        $\mathcal{L}_{MNTP}$ & 35.46 \\ 
        $\mathcal{L}_{SC}$ & 41.17 \\  
        
        \bottomrule
        \\
        \end{tabular}
    \end{center}}\end{minipage}
    }
    \hspace{2em}
    \subfloat[
    Dataset collection.
    \label{tab:ablation_ds}
    ]{
    \hspace{1pt}
    \begin{minipage}{0.24\linewidth}{\begin{center}
    \tablestyle{2pt}{1.0}
        \begin{tabular}{l c}
        \toprule
        \textbf{Setup} & \textbf{Accuracy} \\
        \midrule
        Ours & \textbf{42.38}\\        
        w/o question verification &  41.51   \\  
        w/o sim-based neg. filter & 41.34 \\
        w/o KL-based neg. filter & 40.71 \\
        \bottomrule
        \end{tabular}
    \end{center}}\end{minipage}
    }
    \centering
    \hspace{3em}
    \subfloat[
    Context-aware emb.
    \label{tab:ablation_sentence_embed}
    ]{
    \begin{minipage}{0.24\linewidth}{\begin{center}
    \tablestyle{2pt}{1.0}
    \begin{tabular}{l c}
    \toprule
        \textbf{Setup} & \textbf{Accuracy} \\
        \midrule
        2 negatives & \textbf{42.38} \\
        4 negatives & 41.52 \\
        8 negatives & 41.06 \\ 
    \bottomrule
    \\
    \end{tabular}
    \end{center}}\end{minipage}
    }
}
\caption{Ablation and analysis of different components of CPC, conducted on the SingleDoc subset of LongBench.
}
\label{tab:ablations}
\end{table*}

\subsection{Results}
\subsubsection{Main results.} The main results of our proposed solution on the LongBench and ZeroSCROLLS datasets are presented in Table \ref{tab:main_result_long_context}. Following prior works, we report the results on LongBench on the sub-tasks of a single document, multi-document, summarization, few-shot, synthetic, code, and the average over all the tasks. Following \citet{jiang2023longllmlingua}, the results are reported on 3,000 and 2,000 tokens, which is about 3$\times$ and 5$\times$ compression of the total token length. For ZeroSCROLLS, we present the performance with the same 3,000 and 2,000 tokens. As we observe in this table, our method outperforms the prior works by a considerable margin on LongBench with both 2,000 and 3,000 token constraints. More specifically, on 2,000 tokens, CPC outperforms others by 1.2 points on average and up to 8.3 points on individual tasks. On evaluation with 2,000 tokens, CPC shows a 1.5 points improvement on average. On ZeroSCROLLS, our proposed method also outperforms the previous state-of-the-art by 1.4 points on 3,000 tokens and 0.4 points on 2,000 tokens.

\begin{table*}
    \centering
    \setlength
\tabcolsep{7pt}
    \begin{tabular}{lccccccc}
        \toprule
        {\textbf{Embed.}} & {\textbf{Sin. Doc}} & {\textbf{Mul. Doc}} & {\textbf{Summ.}} & {\textbf{FewShot}} & {\textbf{Synth.}} & {\textbf{Code}} & {\textbf{AVG}}\\ \midrule
        
        MpNet-v2 & 38.4 & 47.5 & 23.5 & 65.1 & 40.5 & 55.5 & 47.2 \\
        LLM2Vec & 38.3 & 45.7 & \textbf{24.7} & 67.5 & 27.0 & 54.3 & 42.9 \\
       \textbf{Ours} & \textbf{42.6} & \textbf{48.6} & 23.7 & \textbf{69.4} & \textbf{52.8} & \textbf{60.0} & \textbf{49.5} \\ 
    \bottomrule        
    \end{tabular}
    \caption{Performance with different sentence encoders.}
    \label{tab:embedding_models}
\end{table*}

\subsubsection{Domain generalization.}
Since CPC effectively removes sentences that do not contain any useful information regarding the question or the prompt, it holds the potential for a wide range of tasks/domains. To this end, we investigate the performance of CPC on new domain-specific benchmarks described earlier in Section \ref{sec:dataset}. The results of these experiments are presented in Table \ref{tab:domain_gen} with a 2,000-token constraint. As we observe from this table, CPC achieves significant improvements over prior methods across all domain-specific tasks. Specifically, CPC outperforms LLMLingua-2 by 6.8, 1.35, 0.93, and 0.26 points on Krapivin-2009, PubMed, MeetingBank, and SummScreen, respectively.

\subsubsection{Larger backbone.}
To evaluate the scalability of our method with a larger model, we compare the performance of CPC against prior works with a larger LLM, GPT-4o, as the backbone. The results of this study are presented in Table \ref{tab:gpt4o}. As we find from this table, CPC consistently outperforms existing methods across all subtasks of the LongBench. On average, CPC outperforms LLMLingua-2 and LongLLMLingua by 12.22 and 21.49 points, respectively.

\subsubsection{Latency evaluation.}
Here, we discuss the latency of our proposed solution and compare it to prior works on prompt compression. We summarize the findings in Table \ref{tab:latency}. All the evaluations are conducted on the same A100 GPU used for training. The reported results are the average processing time of a sample of the LongBench. We report the results for our method in comparison to three prior SOTAs: LLMLingua \cite{jiangllmlingua}, LLMLingua-2 \cite{pan2024llmlingua}, and LongLLMLingua \cite{jiang2023longllmlingua}. As we find from this study, our method is significantly faster than LLMLingua and LongLLMLingua, showing up to $27.5\times$ and $10.93\times$ average and median speedup. While LLMLingua-2 is slightly faster than our solution during inference, its performance is considerably worse than ours in all benchmarks and even worse than LongLLMLingua in most tasks. We also analyze the effect of the length of the original context on the latency of each of the above approaches. We depict these results in Figure \ref{fig:latency}. As we find from the Figure, the time complexity of our approach scales linearly with the growing length of the input context, while having relatively low latency compared to LongLLMLingua. On the other hand, LongLLMLingua has high latency for some of the samples with shorter context due to the dependence on the text structure, while our method is agnostic to the text structure.

\subsubsection{Ablation studies.}
Here we present ablation studies on different components of the proposed method. All the ablation studies are conducted on the LongBench subset SingleDoc consisting of NarrativeQA, Qasper and MultiFieldQa-en datasets. We perform these ablations on the higher compressed ratio with 2,000 tokens. First, we present an ablation study on the loss function used for training the context-aware sentence encoder. As discussed earlier, we train the encoder with the combination of two losses, $\mathcal{L}_{SC}$ and $\mathcal{L}_{MNTP}$. In Table \ref{tab:ablation_loss}, we present the results for removing each of these components. As we find from this study, removing the $\mathcal{L}_{MNTP}$ shows a large drop in performance since $\mathcal{L}_{MNTP}$ helps the model to learn the bidirectional attention that captures the context of the whole sequence. Similarly, removing $\mathcal{L}_{SC}$ also results in a large drop in the performance as the ablated model does not learn the context-aware embeddings that are required to understand the relevance of each sentence to the given context and question.

Next, we ablate the design choices of different components of the dataset collection procedure. We note that the dataset collection has several steps of verification and filtering: question and answer verification, similarity-based negative filtering, and KL-based negative filtering. In Table \ref{tab:ablation_ds}, we present an ablation study by removing each of these steps from our dataset collection pipeline. As we find from this table, removing question verification results in a 0.87 point drop in performance. This is caused by the fact that without question verification, the sentence and positive pairs are generated such that the question is directly fully answerable without the need to consider the context. Next, we find that removing the similarity-based negative filtering and KL-based filtering results in a 1.04 and 1.67 points drop in performance, respectively, showing the importance of these steps in ensuring that the negative does not contain any information regarding the question.

Next, we study the context-aware sentence embedding learning with different numbers of negatives. More specifically, we study 2, 4, and 8 negatives per positive samples in Table \ref{tab:ablation_sentence_embed}. The results indicate that the best context-aware sentence embedding is learned with 2 negatives per positive sample. Adding more negatives results in an increase in the computational cost during training while also showing reduced performance compared to 2 negatives.

Finally, we study the impact of our learned context-aware encoder versus pre-trained off-the-shelf semantic encoders. The results from this study are presented in Table \ref{tab:embedding_models}, comparing our encoder to MpNet-v2 and LLM2Vec. As we find from this table, semantic embeddings are considerably less effective in finding the relevance of context sentences to a question to perform context compression. The average scores for MpNet-v2 and LLM2Vec are 2.3\% and 6.6\% lower than our proposed context-aware sentence encoder.

\subsubsection{Qualitative analysis.}
We present a qualitative analysis of our prompt compression method in Figure \ref{fig:color-map}. Here, the first line shows a question, and the rest of the paragraph is the input context. The intensity of the colour indicates the relevance score of each sentence to the question predicted by our method. As we observe in this example, the sentences that are relevant to the question (`What is the GhostVLAD approach?') have higher scores than sentences that have less relevance.

\section{Conclusion}
In this work, we address the problem of prompt compression by developing a novel sentence-level context-aware compression method. To train our method, we first generated a new dataset of questions, positive, and negative pairs using a pre-trained LLM, followed by several steps of verification and filtering. We then use this dataset to contrastively train our novel context-aware sentence encoder. Detailed experiments demonstrate that our method is considerably faster than the state-of-the-art while showing considerable improvements on benchmark question-answering datasets. Additionally, our method shows strong domain generalization and robust performance with larger encoders. We believe our work will foster research in this promising direction of reducing the inference cost of LLMs.

\bibliography{aaai25}

\end{document}